# A Reconstruction System for Industrial Pipeline Inner Walls Using Panoramic Image Stitching with Endoscopic Imaging

Rui Ma
*Shenzhen International Graduate School, Tsinghua University*
Shenzhen, China
mar25@mails.tsinghua.edu.cn

Yifeng Wang
*Shenzhen International Graduate School, Tsinghua University*
Shenzhen, China
wangyifeng25@163.com

Ziteng Yang
*Shenzhen International Graduate School, Tsinghua University*
Shenzhen, China
yangzt24@mails.tsinghua.edu.cn

Jing Guo
*Shenzhen International Graduate School, Tsinghua University*
Shenzhen, China
guoj24@mails.tsinghua.edu.cn

Naomi Imali Okanda
*Shenzhen International Graduate School, Tsinghua University*
Shenzhen, China
naomiokanda7@gmail.com

Xinghui Li*
*Shenzhen International Graduate School, Tsinghua University*
Shenzhen, China
li.xinghui@sz.tsinghua.edu.cn

***Abstract*—Visual analysis and reconstruction of pipeline inner walls remain challenging in industrial inspection scenarios. This paper presents a dedicated reconstruction system for pipeline inner walls via industrial endoscopes, which is built on panoramic image stitching technology. Equipped with a custom graphical user interface (GUI), the system extracts key frames from endoscope video footage, and integrates polar coordinate transformation with image stitching techniques to unwrap annular video frames of pipeline inner walls into planar panoramic images. Experimental results demonstrate that the proposed method enables efficient processing of industrial endoscope videos, and the generated panoramic stitched images preserve all detailed features of pipeline inner walls in their entirety. This provides intuitive and accurate visual support for defect detection and condition assessment of pipeline inner walls. In comparison with the traditional frame-by-frame video review method, the proposed approach significantly elevates the efficiency of pipeline inner wall reconstruction and exhibits considerable engineering application value.**

## I. Introduction

Industrial pipelines are extensively applied in the petrochemical, rail transit, automotive manufacturing and other industries. Surface defects such as scratches, cracks and pits on their inner walls directly compromise the operational safety and reliability of pipelines[1]. As a core tool for non-destructive detection of pipeline inner walls, industrial endoscopes can acquire video data of inner walls without damaging the pipeline structure[2]. However, conventional methods for reviewing endoscope video footage suffer from prominent limitations: on the one hand, video frames are displayed from an annular perspective, failing to intuitively reflect the overall condition of pipeline inner walls; on the other hand, frame-by-frame video analysis is time-consuming and labor-intensive, making it difficult to quickly locate the position and scope of defects[3].

Panoramic image stitching technology offers an effective solution to this problem. By estimating the geometric correspondence among adjacent or overlapping views and blending them into a seamless mosaic, image stitching can fuse a sequence of local perspective images into a complete panoramic image, thereby realizing holistic visual reconstruction of pipeline inner walls [4], [5]. This technique has been widely studied in general panoramic imaging and endoscopic image-sequence reconstruction [6]. In recent years, several studies have further introduced panoramic unfolding and image stitching into pipe or pipeline inspection, including pipe inner-surface panoramic generation, non-centralized pipe image stitching, and endoscopic reconstruction of pipeline inner walls[7]. However, existing stitching algorithms tailored for industrial endoscope videos still suffer from limited adaptability to annular-view distortion, illumination variation, probe eccentricity, and unstable camera motion, making it difficult to satisfy the rapid detection requirements of engineering sites [8].

Panoramic image stitching technology offers an effective solution to this problem. By estimating geometric correspondences among adjacent or overlapping views and blending them into a seamless mosaic, this technology can fuse a sequence of local perspective images into a complete panoramic image, thereby realizing holistic visual reconstruction of pipeline inner walls [9]. At present, some studies have applied image stitching and panoramic reconstruction techniques to pipeline or deep-hole inner-wall inspection. However, existing stitching algorithms designed for industrial endoscope videos still suffer from limited adaptability, complex calibration procedures, and insufficient real-time performance, making them difficult to meet the rapid detection demands of engineering sites.

Many scholars have conducted relevant research on image stitching and visual reconstruction for pipeline inner walls. Sheng et al. [10], [11] developed an imaging model and a distortion-correction system for deep-hole endoscopic images, but the system is relatively complex and cannot fully satisfy real-time processing requirements. Wei et al. [12] proposed a polar-coordinate parameterization method for surround-view perception, which provides useful inspiration for annular image representation in pipeline scenarios. Liu [13] adopted a template-matching approach to locate the center of endoscopic images and extract annular regions based on the identified center; however, this method mainly focuses on single-frame image processing and cannot be directly applied to continuous endoscopic video data. In addition, real-time video stitching methods have been investigated in general vision scenarios [14], but their direct application to industrial endoscope videos remains challenging due to illumination fluctuation, low-texture regions, radial distortion, and irregular camera motion inside pipelines.

Building on the existing research, this paper optimizes the strategy for video key frame extraction and the parameters for polar coordinate unwrapping, and designs a visual interactive interface to achieve fully automated processing from video input to panoramic reconstruction. The core stitching algorithms are encapsulated in a GUI, which enables visual adjustment of video processing parameters, real-time preview of stitching effects, one-click saving of results, and ultimately completes the panoramic reconstruction of pipeline inner walls. The system can convert annular endoscope video frames into planar panoramic images, intuitively presenting the full view of pipeline inner walls; it lowers the technical threshold for operators and improves the efficiency of pipeline detection. Additionally, it provides complete image data support for the quantitative analysis of pipeline defects.

## II. METHODOLOGY

The stitching algorithm for pipeline inner wall images from industrial endoscopes, based on panoramic image stitching and proposed in this paper, mainly consists of three modules: a video key frame extraction algorithm, an annular image unwrapping algorithm and an image stitching algorithm, as shown in Fig. 1.

### *A. Video Key Frame Extraction*

Key frames are extracted from video files captured by industrial endoscopes to balance processing efficiency and reconstruction accuracy. Key-frame extraction is widely used in video summarization and video analysis to reduce redundant frames while preserving representative visual information [15], [16]. Let $N_{total}$ denote the total number of video frames, $N_{interval}$ is the frame interval for key frame extraction, and $M$ is the total number of extracted key frames. The original video frame sequence is defined as $\{F_1,F_2,F_3,...,F_{N_{total}}\}$, and the extracted key frame sequence is $\{F'_1,F'_2,F'_3,...,F'_M\}$.

The *k-th* key frame in the extracted sequence is calculated as follows:

$$F'_k=F_{k\times N_{interval}} \tag{1}$$

The total number of extracted key frames is determined by:

$$M=\lfloor \frac{N_{tatal}}{N_{interval}} \rfloor \tag{2}$$

where $\lfloor \cdot \rfloor$ represents the floor operation.

### *B. Annular Image Unwrapping*

For annular images of pipeline inner walls captured by industrial endoscopes, polar coordinate transformation is employed to unwrap the annular images into rectangular planar images. Similar polar-coordinate or coordinate-mapping strategies have been used for borehole-wall and inner-surface panoramic image generation [17], [18]. Let the polar coordinates of a pixel point $P$ in the original annular image be $(r_p,\theta_p)$, with its corresponding Cartesian coordinates being $(x_p,y_p)$. The coordinate transformation is expressed as:

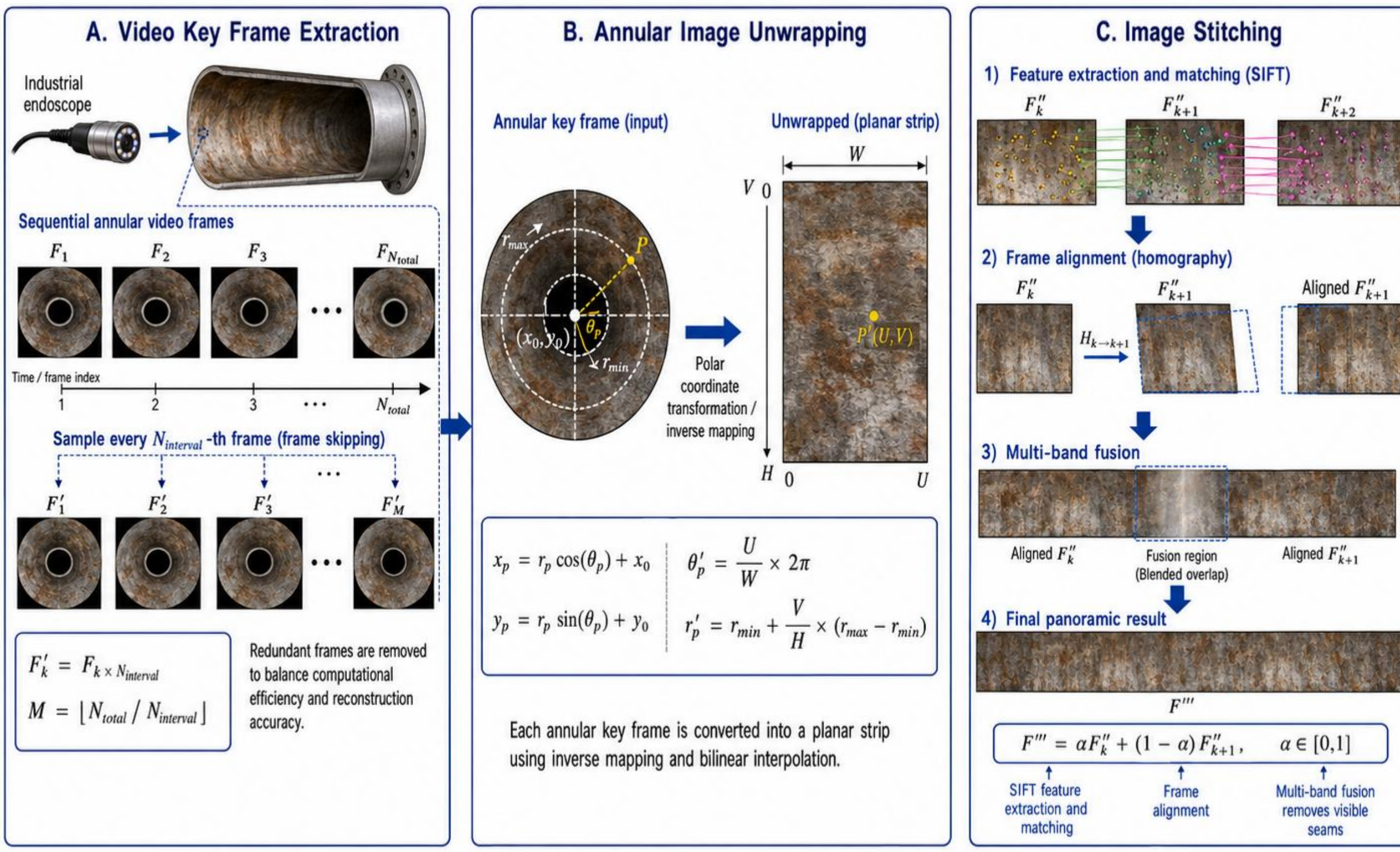

Fig. 1. Panoramic image Stitching method

$$x_p=r_p\times cos(\theta_p)+x_0 \tag{3}$$

$$y_p=r_p\times sin(\theta_p)+y_0 \tag{4}$$

where $(x_0,y_0)$ is the center coordinate of the annular pipeline inner wall image, $\theta_p$ is the polar angle of pixel point $P$, and $r_p$ is the Euclidean distance from pixel point $P$ to the image center, satisfying the constraint $r_{min}\leq r_p\leq r_{max}$.

To realize the unwrapping from annular to planar images, inverse mapping is used to calculate the pixel values of the unwrapped planar image. Let the pixel coordinates of a point $P$ in the unwrapped image be $(U_p,V_p)$, with the corresponding polar coordinates being $(r'_p,\theta'_p)$. The inverse transformation is defined as:

$$\theta'_p=\frac{U}{W}\times 2\pi \tag{5}$$

$$r'_p=r_{min}+\frac{V}{H}\times(r_{max}-r_{min}) \tag{6}$$

where $U$ is the horizontal coordinate of the unwrapped image with the range $0\leq U\leq W$, $V$ is the vertical coordinate with $0\leq V\leq H$, $W$ is the width of the unwrapped image, and $H$ is the height of the unwrapped image.

By combining equations (3)-(6), the corresponding coordinates $(x'_p,y'_p)$ in the original annular image are obtained for each pixel $(U'_p,V'_p)$ in the unwrapped planar image. Bilinear interpolation is then applied to calculate the pixel values $F''(U'_p,V'_p)$ of the polar coordinate-unwrapped image, yielding the key frame sequence after polar coordinate unwrapping: $\{F''_1,F''_2,F''_3,...,F''_M\}$.

### C. Image Stitching

The key frame sequence $\{F''_1,F''_2,F''_3,...,F''_M\}$ after polar coordinate unwrapping is stitched together via feature point matching. The Scale-Invariant Feature Transform (SIFT) algorithm is used to extract feature points from adjacent key frames because it provides local features that are robust to scale and rotation variations [19]. Let the feature point sets of the *k-th* and *(k+1)-th* frames be $P_k=\{(x_{k,1},y_{k,1}),(x_{k,2},y_{k,2}),...,(x_{k,n},y_{k,n})\}$ and $P_{k+1}=\{(x_{k+1,1},y_{k+1,1}),(x_{k+1,2},y_{k+1,2}),...,(x_{k+1,n},y_{k+1,n})\}$ respectively; Euclidean clustering is then utilized to match the corresponding feature points between the two frames.

A multi-band fusion algorithm is adopted to eliminate visible stitching seams between adjacent frames. Let $F'''$ represents the final stitched image; the pixel values in the fusion region of the *k-th* frame are calculated as:

$$F'''=\alpha F''_k+(1-\alpha)F''_{k+1} \tag{7}$$

where $\alpha$ is the fusion weight with a value range of $\alpha\in[0,1]$, and it is determined by linear interpolation based on the distance from each pixel to the stitching seam.

## III. Experiment and Discussion

The experimental validation was conducted using a metal pipeline specimen with an axial depth of 600 mm. Image data were acquired by an industrial endoscope inserted into the pipeline, and the endoscope was translated along the pipeline axis at a constant speed of 30 mm/s. The video acquisition device for the pipeline inner wall is as shown in Fig. 2. The captured video had a spatial resolution of $1920\times 1080$ pixels and a frame rate of 30 frames/s, which provided sufficient temporal and spatial information for subsequent key-frame extraction and panoramic reconstruction.

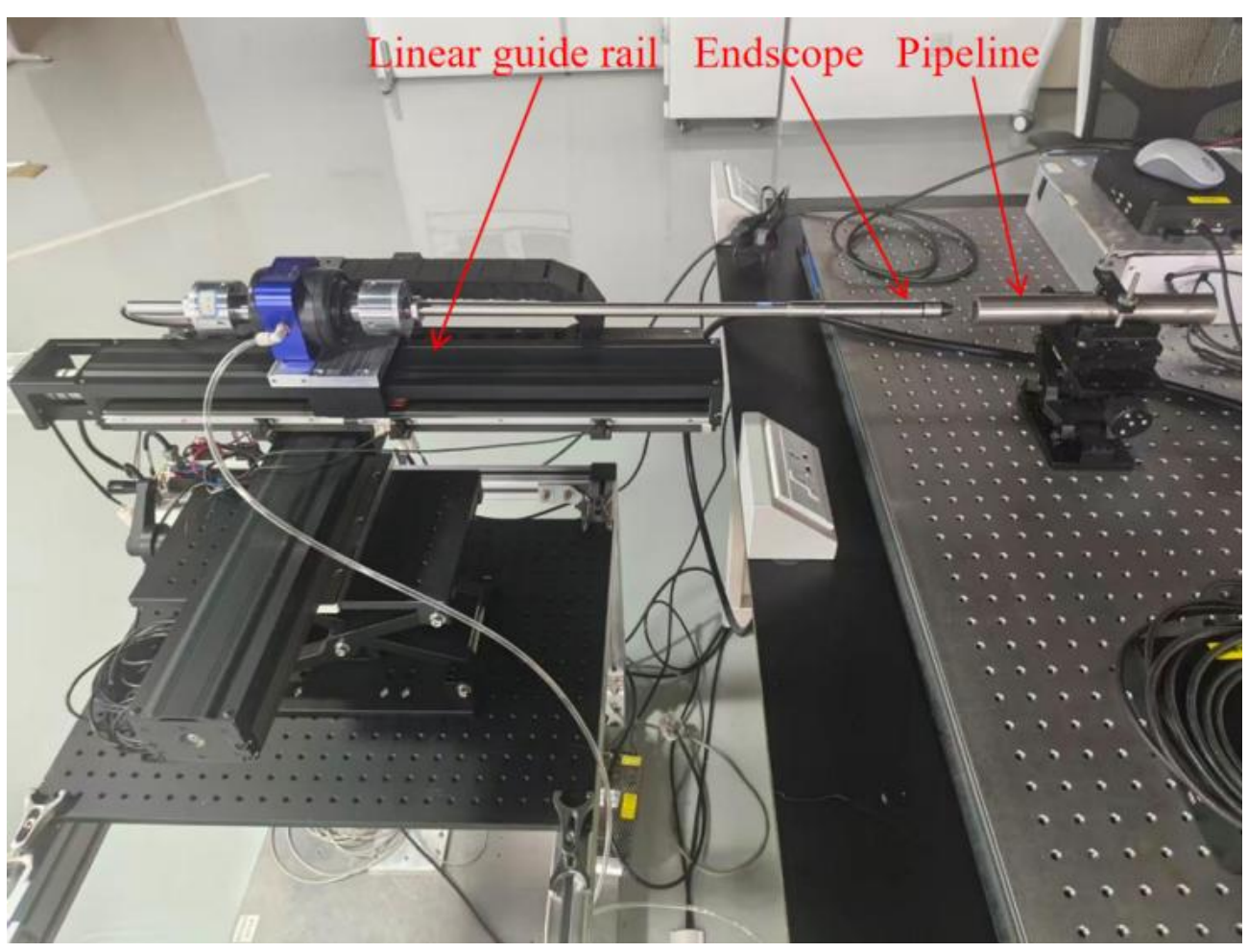

Fig. 2. Video acquisition device for the pipeline inner wall

All video processing and image reconstruction procedures were implemented on an industrial computer equipped with an Intel Core i7-12700H processor and 16 GB RAM. The software framework was developed using Python 3.9 and OpenCV 4.8.0.

In the experiment, the main geometric and stitching parameters were normally set as follows: $x_0=540$, $y_0=960$, $W=500$, $H=250$, $r_{min}=150$, $r_{max}=500$, and $N_{interval}=10$. Here, $(x_0,y_0)$ denotes the center position used for annular image unwrapping, $r_{min}$ and $r_{max}$ define the inner and outer radii of the valid annular region, $W$ and $H$ represent the width and height of the unwrapped image strip, respectively, and $N_{interval}$ controls the interval for key-frame selection during video processing.

With the above parameter configuration, the proposed system successfully transformed the annular endoscopic video frames of the pipeline inner wall into a planar panoramic representation. The reconstructed panoramic image provides a continuous 360° visualization of the inner wall surface along the axial direction of the pipeline. Compared with the original annular frames, the panoramic image offers a more intuitive representation of the spatial distribution of surface features and defects. In particular, defects such as corrosion regions and scratches can be clearly observed, and their circumferential positions, axial extension, and relative lengths can be more conveniently identified from the reconstructed result.

The stitching result demonstrates that the proposed method can effectively preserve the structural details of the pipeline inner wall while maintaining good visual continuity between adjacent image strips. No obvious stitching seams or severe geometric discontinuities are observed in the final panoramic image. This indicates that the combination of key-frame extraction, annular image unwrapping, and sequential image stitching is suitable for inner-wall inspection scenarios using industrial endoscopic video. The final stitching result is shown in Fig. 3.

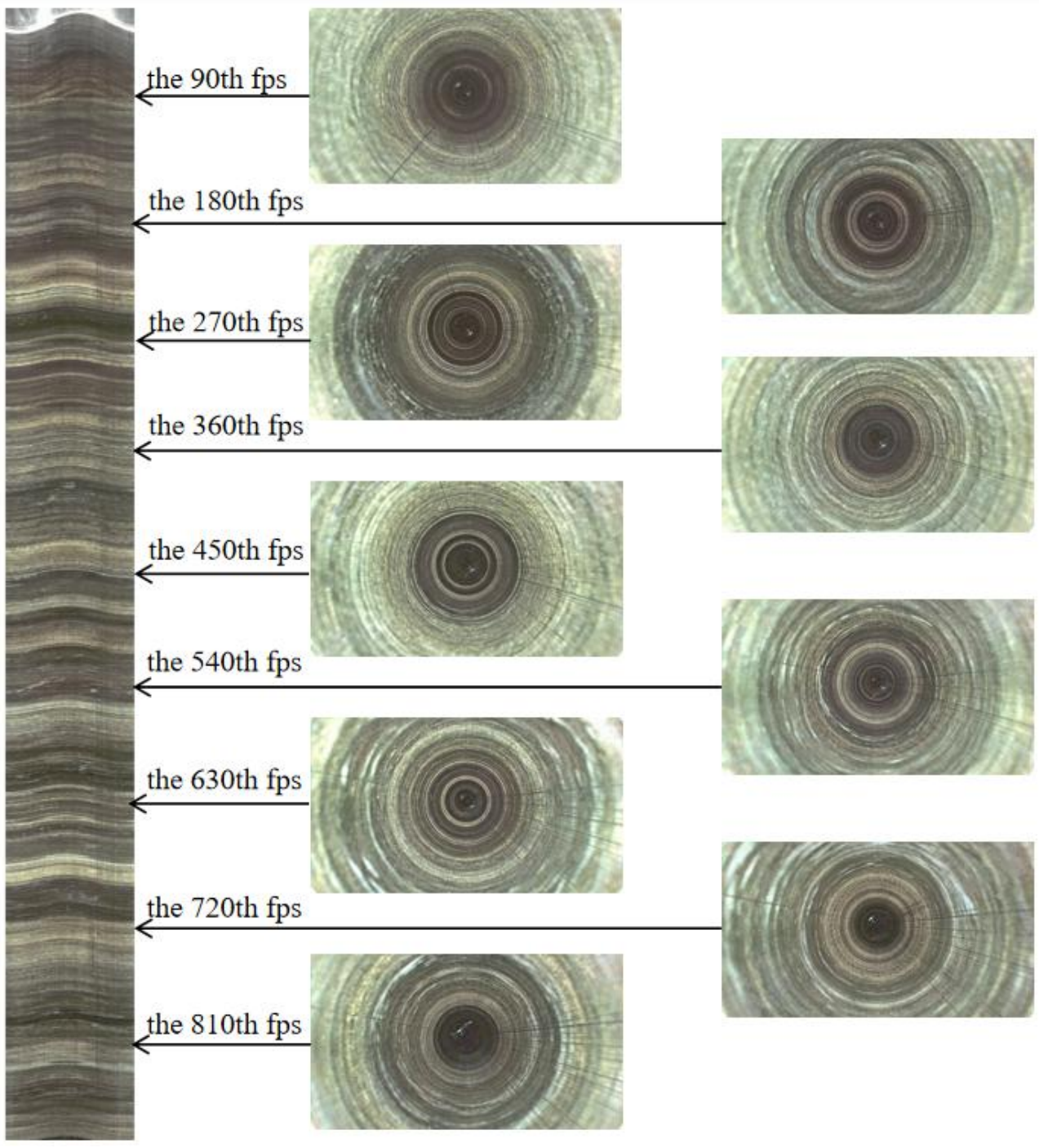

Fig. 3. Panoramic stitched image of industrial pipeline inner wall via endoscope

### A. *Parameter Sensitivity Analysis*

a) Influence of frame interval: As the frame interval increased from 1 to 20, the total video processing time decreased from 24 seconds to 5 seconds, with no significant degradation in stitching accuracy. However, when the frame interval exceeded 30, obvious discontinuities and artifacts appeared in the stitched panoramic image. Therefore, the optimal value range for the frame interval is determined to be 5~20.

b) Influence of center coordinates: When the deviation of the manually set center coordinates exceeded ±50 pixels from the actual center of the annular image, the unwrapped planar image exhibited obvious stretching or compression distortions. Thus, the center coordinates need to be fine-tuned according to the actual center position of the video frames.

c) Influence of radius range: An excessively small minimum radius ( $r_{min}$ ) would introduce background noise from the area outside the pipeline into the unwrapped image, while an excessively large maximum radius ( $r_{max}$ ) would result in blurring at the edges of the unwrapped image. Hence, the radius range parameters should be adjusted according to the actual inner diameter of the detected pipeline

### B. *Performance Analysis*

The system took approximately 7 seconds to process a 20-second endoscope video with a frame interval set to 10, which represents a significant efficiency improvement compared with the traditional manual frame-by-frame analysis method. The resolution of the stitched panoramic image can reach 4500×500 pixels, which fully meets the detailed requirements for pipeline inner wall defect detection. In addition, the system's key frame saving function enables batch storage of the extracted effective frames, facilitating subsequent secondary analysis and defect quantification by inspection personnel.

## IV. CONCLUSION

This paper presents an industrial endoscope pipeline inner wall reconstruction system based on panoramic image stitching technology. By designing a visual interactive interface and an adaptive stitching algorithm, the system achieves efficient conversion from annular pipeline inner wall video frames to high-quality planar panoramic stitched images. The system integrates three core algorithms: video key frame extraction, polar coordinate transformation and image stitching, and its adjustable parameters allow it to adapt to the processing requirements of endoscope video footage captured under different industrial scenarios and shooting conditions. Experimental results verify that the system can intuitively and accurately reconstruct the full view of pipeline inner walls, significantly improve the efficiency of industrial pipeline detection, and effectively solve the problems of unintuitive viewing and low analysis efficiency associated with traditional endoscope video review methods.

The graphical user interface of the system reduces the technical operation threshold, making it easy for on-site inspection personnel to master; its flexibly adjustable parameters enable adaptation to endoscope video footage of pipelines with different diameters and captured under various shooting conditions, demonstrating excellent engineering practicability. In future research, we will focus on optimizing the system for processing low-quality endoscope video footage (e.g., blurry or noisy images) and improving its adaptability to curved pipelines. This will further expand the application scenarios of the proposed method and provide more comprehensive technical support for the non-destructive detection of industrial pipelines.

## ACKNOWLEDGMENT

The authors would like to thank all the people who have contributed to this work. We are deeply grateful to our advisors for their valuable guidance, insightful suggestions, and continuous support throughout the research and paper writing. We also thank all colleagues and lab mates for helpful discussions and assistance in experiments. In addition, we appreciate the anonymous reviewers for their constructive comments to improve this paper.